\documentclass[letterpaper]{article} 
\usepackage{aaai25}
\nocopyright
\usepackage{times}  
\usepackage{helvet}  
\usepackage{courier}  
\usepackage[hyphens]{url}  
\usepackage{graphicx} 
\usepackage{natbib}  
\usepackage{caption} 
\usepackage{algorithm}
\usepackage{algorithmic}

\usepackage{multirow}
\usepackage{graphicx}
\usepackage{booktabs}
\usepackage{caption}
\usepackage{array}
\usepackage{makecell}
\usepackage{amsmath} 
\usepackage{amssymb} 
\usepackage{xcolor}

\usepackage{newfloat}
\usepackage{listings}
\DeclareCaptionStyle{ruled}{labelfont=normalfont,labelsep=colon,strut=off} 
\floatstyle{ruled}
\newfloat{listing}{tb}{lst}{}
\floatname{listing}{Listing}
\title{\textsc{Learning from Gene Names, Expression Values and Images}: Contrastive Masked Text-Image Pretraining for Spatial Transcriptomics Representation Learning}

\author{
  Jiahe Qian\textsuperscript{\rm 1,4},
  Yaoyu Fang\textsuperscript{\rm 1},
  Ziqiao Weng\textsuperscript{\rm 1},
  Xinkun Wang\textsuperscript{\rm 2},
  Lee A. Cooper\textsuperscript{\rm 3},
  Bo Zhou\textsuperscript{\rm 1}\thanks{Corresponding author.}
}
\affiliations{
  \textsuperscript{\rm 1}Department of Radiology, Northwestern University, Chicago, IL 60611, USA\\
  \textsuperscript{\rm 2}Department of Cell and Developmental Biology, Northwestern University, Chicago, IL 60611, USA\\
  \textsuperscript{\rm 3}Department of Pathology, Northwestern University, Chicago, IL 60611, USA\\
  \textsuperscript{\rm 4}Institute of Automation, Chinese Academy of Sciences, Beijing 100190, China\\
  jiahe.qian@northwestern.edu,\;
  bo.zhou@northwestern.edu
}

\usepackage{bibentry}

\begin{document}

\maketitle

\begin{abstract}
Spatial transcriptomics aims to connect high-resolution histology images with spatially resolved gene expression. To achieve better performance on downstream tasks such as gene expression prediction, large-scale pre-training is required to obtain generalisable representations that can bridge histology and transcriptomics across tissues, protocols, and laboratories.  Existing cross-modal pre-training approaches for spatial transcriptomics rely on either gene names or expression values in isolation, which strips the gene branch of essential semantics and breaks the association between each gene and its quantitative magnitude. In addition, by restricting supervision to image-text alignment, these methods ignore intrinsic visual cues that are critical for learning robust image features.  We present \textit{\textbf{CoMTIP}}, the first \textit{\textbf{Co}}ntrastive \textit{\textbf{M}}asked \textit{\textbf{T}}ext-\textit{\textbf{I}}mage \textit{\textbf{P}}retraining framework that jointly learns from images, gene names, and expression values while capturing fine-grained visual context for spatial transcriptomics.  The vision branch uses Masked Feature Modeling to reconstruct occluded patches and learn context-aware image embeddings. The text branch applies a scalable Gene-Text Encoder that processes all gene sentences in parallel, enriches each gene and its numerical value with dedicated embeddings, and employs Pair-aware Adversarial Training (PAAT) to preserve correct gene–value associations. Image and text representations are aligned in a shared InfoNCE-optimised space.  Experiments on public spatial transcriptomics datasets show that CoMTIP not only surpasses previous methods on diverse downstream tasks but also achieves zero‑shot gene expression prediction, a capability that existing approaches do not provide. Code and pretrained model will be released.
\end{abstract}

\section{Introduction}


Spatial transcriptomics enriches histopathology by providing gene-expression profiles that are localised within tissue sections.  This fusion of morphologic context and molecular readout has become a pivotal paradigm for dissecting cellular heterogeneity in cancer progression, developmental biology, and precision medicine \cite{vickovic2019high, chen2022spatiotemporal, gong2024spatial, staahl2016visualization, rodriques2019slide, moncada2020integrating, stickels2021highly, kuppe2022spatial}.  Despite its promise, most computational studies rely on models trained on a single dataset, which restricts scalability and hampers generalisation across platforms and laboratories \cite{he2020integrating, chung2024accurate, liu2024ressat, zheng2024digital, xie2023spatially, song2024predicting, schmauch2020deep, mondol2023hist2rna, zhu2025dusted, ganguly2025merge}. Building pre‑training models that can reason jointly over large collections of histology images and genome‑scale expression profiles would enable comprehensive comparative atlases, accelerate biomarker discovery, and ultimately support data‑driven therapeutic decision making.  

\begin{figure*}
    \centering
    \includegraphics[width=0.98\linewidth]{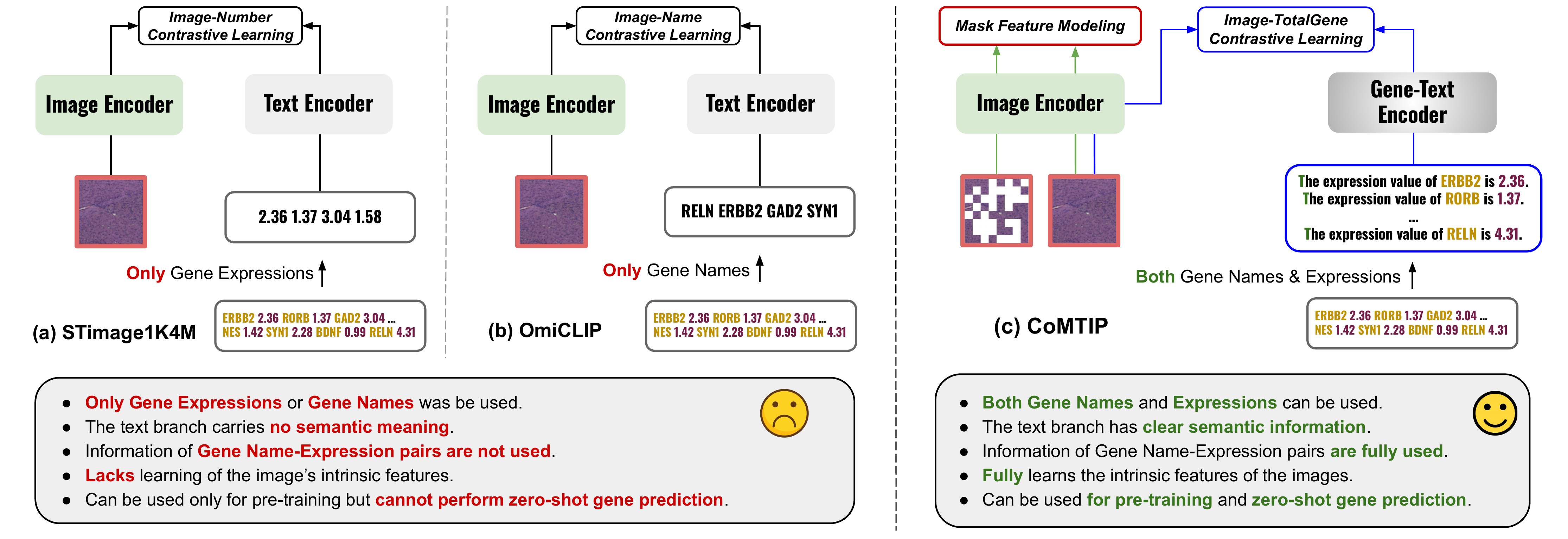}
    \caption{Comparison of spatial transcriptomics pre-training paradigms.  (a) STimage‑1K4M \cite{chen2024stimage} encodes only gene expressions and lacks robust visual feature learning.  (b) OmiCLIP \cite{chen2025visual} focuses on gene names alone and exhibits similar restrictions.  (c) Our CoMTIP couples Masked-Feature Modeling with a Gene-Text Encoder and jointly aligns histology images with gene name–value pairs, yielding semantically complete representations together with robust visual features. Moreover, it also enable zero-shot gene prediction}
    \label{fig:3}
\end{figure*}

Recent vision-language pre-training studies show that contrastive objectives can bridge heterogeneous modalities \cite{radford2021learning, zuo2024plip, wang2022medclip, jia2021scaling, li2021align, li2022blip, wang2021simvlm}. Yet current cross-modal pre-training models for spatial transcriptomics continue to exhibit two major limitations \cite{chen2024stimage, chen2025visual}, as illustrated in Figure~\ref{fig:3}. First, existing approaches encode only one part of the transcriptomic description, which removes complementary semantics and breaks the link between each gene and its magnitude.  For example, STimage\textendash1K4M \cite{chen2024stimage} retains only numerical expression vectors, whereas OmiCLIP \cite{chen2025visual} retains only gene names. Second, the visual encoders in these methods are trained solely by image-text alignment and lack explicit mechanisms for modeling within-slide variation. Without dedicated feature reconstruction or context modeling, their embeddings capture sparse global cues but may have the risk of missing fine-grained structures that are critical for robust vision representation.

We tackle these issues with CoMTIP, a unified cross-modal pre-training framework that couples masked-feature modeling on the vision side with pair-aware contrastive learning on the text side. Masked-feature modeling hides about half of the image patches and drives the network to reconcile visible and occluded regions through a dual-distillation objective embedding the reconstruction signal directly into the contrastive cycle \cite{he2022masked, xie2022simmim}. In the text branch, the proposed Gene-text Encoder hosts multiple Gene-Sentence Encoders that handle individual gene sentences, augment each gene name and its expression value with dedicated embeddings. By processing sentences independently, all gene name–expression pairs can be ingested simultaneously.  Their features are globally pooled, and during training Pair-Aware Adversarial Training (PAAT) presents randomly mismatched sentences to guide the encoder toward pair-aware representations. Projection heads map both modalities into a shared space optimized with the InfoNCE objective \cite{he2020momentum}. The resulting model exhibits strong robustness and supports zero-shot gene prediction as well as other spatial transcriptomics tasks. Our main contributions are threefold:


\begin{itemize}
    \item CoMTIP offers the first genome-scale pre-training model that aligns whole-slide imagery with gene identities and expression magnitudes in a single latent space, thereby supporting zero-shot molecular inference and broad transfer across spatial transcriptomics tasks.
    \item Its vision branch introduces Masked‑Feature Modeling, a feature‑level strategy that inserts learnable queries for hidden patches and forces the encoder to reconstruct their features, thereby producing context‑aware and noise‑robust image embeddings that go beyond pixel‑wise MAE.
    \item A scalable Gene-Text Encoder attaches dedicated embeddings to every gene name and numerical value, paired with PAAT to preserve accurate gene–value semantics across thousands of sentences and enhances downstream predictive accuracy.
\end{itemize}

\section{Method}

In this section, we first introduce the overall pipeline of the CoMTIP. Then, we elaborate on our key designs of (1) Masked-Feature Modeling for image encoder context-aware learning, and (2) Gene-Text Encoder with PAAT for securing correct gene–value pairing. In the final subsection, we detail the complete loss formulation that unifies contrastive alignment, reconstruction, and pairing regularisation.


\subsection{Overall Pipeline of CoMTIP}

\begin{figure*}[t]  
    \centering
    \includegraphics[width=0.925\linewidth]{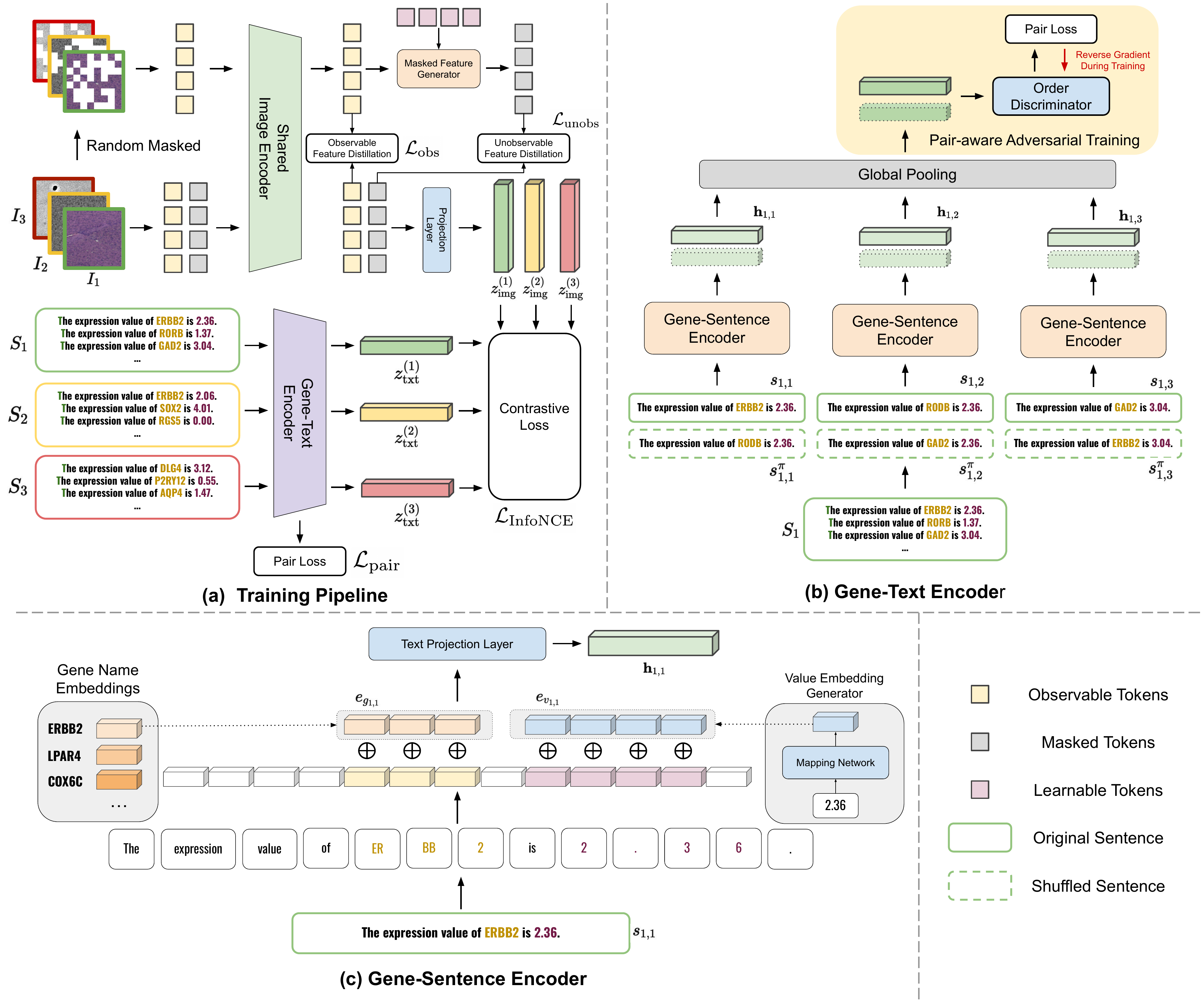}
    \caption{Overview of CoMTIP. (a) Training applies masked feature modeling and a contrastive loss to align image and text embeddings. (b) Gene-Text Encoder averages outputs from parallel Gene-Sentence Encoders and employs PAAT to preserve correct gene–value pairing. (c) Each Gene-Sentence Encoder enriches gene and value tokens with dedicated embeddings before projection to the shared latent space.}
    \label{fig:4}
\end{figure*}



Figure~\ref{fig:4}\,(a) presents the end-to-end workflow of CoMTIP.  The framework receives a mini-batch of whole-slide histology images \(\mathcal{I}=\{I_{k}\}_{k=1}^{m}\) together with corresponding collections of gene–expression sentences \(\mathcal{S}=\{S_{k}\}_{k=1}^{m}\), where each \(S_{k}=\{s_{k,i}\}_{i=1}^{N}\) is obtained by converting every gene–expression pair into a sentence of the form “The expression value of \textit{gene} is \textit{value}.” and \(m\) denotes the batch size.  

First, the vision branch processes each whole‑slide image \(I_{k}\) through a ViT‑B/32 backbone \(E_{\mathrm{img}}\). Every image \(I_{k}\) passes through a shared vision backbone \(E_{\mathrm{img}}\) that produces patch-level features.  Masked-Feature Modeling randomly hides approximately half of these patches and forwards both complete and masked tensors through a Mask Feature Generator.  This operation compels the backbone to infer missing context and yields robust visual embeddings \(z_{\mathrm{img}}^{(k)}\).

Second, the textual branch feeds the sentence collections in \(\mathcal{S}\) into a Gene-Text encoder \(E_{\mathrm{txt}}\) that contains $i$ parallel Gene-Sentence Encoders.  Each branch processes one sentence, enriching the gene token with a dedicated gene-name embedding and the value token with a value embedding.  The resulting sentence features are averaged to obtain a sample-level vector \(z_{\mathrm{txt}}^{(k)}\).  During training PAAT teaches the encoder to distinguish sentences that contain correct gene–value pairs from sentences whose pairs are randomly shuffled and therefore drives the model to preserve only the cues that encode each true association.

Two linear projection heads \(P_{\mathrm{img}}\) and \(P_{\mathrm{txt}}\) map \(z_{\mathrm{img}}^{(k)}\) and \(z_{\mathrm{txt}}^{(k)}\) into a unified latent space. It offers a powerful initialization for a broad range of spatial transcriptomics downstream applications.


\subsection{Mask Feature Modeling}

Mask Feature Modeling establishes two parallel image branches that share a single encoder but accept different token sets.  Each whole-slide image \(I_{k}\) is tessellated into \(P\) non-overlapping patches whose embeddings constitute the set \(T_{k}=\{\mathbf{x}_{k,i}\}_{i=1}^{P}\) with \(\mathbf{x}_{k,i}\in\mathbb{R}^{d}\).  A binary random mask removes a proportion \(r \in (0,1)\) of tokens to produce an observable subset \(\mathcal{O}_{k}\subset T_{k}\) and an unobservable subset \(\mathcal{U}_{k}=T_{k}\setminus\mathcal{O}_{k}\).  We use \(r=0.5\) in all experiments so exactly half of the spatial context is hidden from the masked branch. The full branch forwards the complete sequence through the shared image encoder
\[
\mathbf{Z}_{k}=E_{\mathrm{img}}\bigl(T_{k}\bigr)\in\mathbb{R}^{P\times d},
\]
and collects features for every patch.  In parallel, the masked branch processes only observable tokens,
\[
\mathbf{H}_{\mathcal{O},k}=E_{\mathrm{img}}\bigl(\{\mathbf{x}_{k,i}\mid i\in\mathcal{O}_{k}\}\bigr)\in\mathbb{R}^{|\mathcal{O}_{k}|\times d},
\]
thereby forcing the encoder to rely on partial visual evidence.

Each missing position \(j\in\mathcal{U}_{k}\) is paired with a learnable query vector \(\mathbf{q}_{k,j}\in\mathbb{R}^{d}\).  All queries and the observable features are fed into the Mask Feature Generator, a single cross-attention layer that reconstructs the deleted information,
\begin{equation}
\widehat{\mathbf{z}}_{k,j}= \mathrm{Attn}\!\left(W_{Q}\mathbf{q}_{k,j},\, W_{K}\mathbf{H}_{\mathcal{O},k},\, W_{V}\mathbf{H}_{\mathcal{O},k}\right),
\end{equation}
where \(W_{Q}, W_{K}, W_{V}\in\mathbb{R}^{d\times d}\) are projection matrices and \(\widehat{\mathbf{z}}_{k,j}\) is the predicted feature of patch \(j\).  Scaled dot-product attention is used so that the query attends to all visible patches and aggregates their information weighted by relevance.

Information transfer from the full branch to the masked branch is enforced through two mean-squared-error terms.  For observable patches the encoder must preserve visible content
\begin{equation}
\mathcal{L}_{\mathrm{obs}}=\frac{1}{|\mathcal{O}_{k}|}\sum_{i\in\mathcal{O}_{k}}\|\mathbf{Z}_{k,i}-\mathbf{H}_{\mathcal{O},k,i}\|_{2}^{2},
\end{equation}
and for unobservable patches it must accurately infer hidden features
\begin{equation}
\mathcal{L}_{\mathrm{unobs}}=\frac{1}{|\mathcal{U}_{k}|}\sum_{j\in\mathcal{U}_{k}}\|\mathbf{Z}_{k,j}-\widehat{\mathbf{z}}_{k,j}\|_{2}^{2}.
\end{equation}
These objectives encourage spatial coherence and strengthen contextual reasoning inside the backbone.

After reconstruction the model derives its image representation exclusively from the full branch.  Teacher patch features are average pooled and passed through a learnable projection layer,
\begin{equation}
z_{\mathrm{img}}^{(k)} = P_{\mathrm{img}}\Bigl(\frac{1}{P}\sum_{i=1}^{P}\mathbf{Z}_{k,i}\Bigr)\in\mathbb{R}^{d},
\end{equation}
where \(P_{\mathrm{img}}\) reduces the pooled vector to the contrastive embedding dimension.  This design markedly improves robustness and delivers reliable visual features for subsequent cross-modal alignment.

\subsection{Gene-Text Encoder}

The Gene-Text Encoder contains \(N\) parallel Gene-Sentence Encoders \(\{E_{\mathrm{gs}}^{(i)}\}_{i=1}^{N}\), each following the architecture of the CLIP text encoder.  Given \(N\) sentences \(S_{k}=\{s_{k,i}\}_{i=1}^{N}\), every gene–expression pair is tokenised and routed to its corresponding Gene-Sentence Encoder. Inside each encoder, the gene token obtains a learnable gene-name embedding
\[
e_{g_{k,i}}\in\mathbb{R}^{d},
\]
fetched from a table that assigns a unique vector to every gene symbol.  The numerical expression magnitude \(v_{k,i}\in\mathbb{R}\) is converted into a value embedding through
\begin{equation}
e_{v_{k,i}} = W_{v}v_{k,i} + \mathbf{b}_{v}\in\mathbb{R}^{d},
\end{equation}
where \(W_{v}\in\mathbb{R}^{d\times1}\) and \(\mathbf{b}_{v}\in\mathbb{R}^{d}\).  Let \(\mathbf{w}_{g_{k,i}}\) and \(\mathbf{w}_{v_{k,i}}\) denote the base embeddings of the gene and value tokens.  They are updated by
\begin{equation}
\widetilde{\mathbf{w}}_{g_{k,i}}=\mathbf{w}_{g_{k,i}}+e_{g_{k,i}}, \qquad
\widetilde{\mathbf{w}}_{v_{k,i}}=\mathbf{w}_{v_{k,i}}+e_{v_{k,i}}.
\end{equation}
Replacing the original tokens with \(\widetilde{\mathbf{w}}_{g_{k,i}}\) and \(\widetilde{\mathbf{w}}_{v_{k,i}}\) forms the modified sentence \(\widetilde{s}_{k,i}\).  
This sentence is then passed through the projection layer \(P_{\mathrm{sent}}\), yielding the sentence embedding
\[
\mathbf{h}_{k,i}=P_{\mathrm{sent}}(\widetilde{s}_{k,i})\in\mathbb{R}^{d}.
\]
Averaging all sentence embeddings in \(S_{k}\) produces the sample-level textual vector
\begin{equation}
z_{\mathrm{txt}}^{(k)}=\frac{1}{N}\sum_{i=1}^{N}\mathbf{h}_{k,i}\in\mathbb{R}^{d}.
\end{equation}



\subsection{Pair-Aware Adversarial Training}

To make \(z_{\mathrm{txt}}^{(k)}\) faithfully encode the correspondence between each gene and its value, we pair the gene‑text encoder with an adversarial objective. For every sample the sentence set \(S_{k}\) is duplicated into a shuffled counterpart \(S_{k}^{\pi}\) by randomly permuting expression values across gene symbols, and the encoder yields the pooled representations \(z_{\mathrm{txt}}^{(k)}\) and \(z_{\mathrm{txt}}^{(k)\pi}\), respectively.  
A discriminator \(D\) is trained to reduce the Wasserstein distance between the two distributions,
\begin{equation}
\mathcal{L}_{\mathrm{pair}}
   =\mathbb{E}_{S_{k}}\!\bigl[D\!\left(z_{\mathrm{txt}}^{(k)}\right)\bigr]
   -\mathbb{E}_{S_{k}^{\pi}}\!\bigl[D\!\left(z_{\mathrm{txt}}^{(k)\pi}\right)\bigr],
\end{equation}
so minimising \(\mathcal{L}_{\mathrm{pair}}\) prompts \(D\) to push the scores of matched and mismatched examples closer.

A gradient‑reversal layer \(\mathcal{R}\) is inserted between the encoder and the discriminator.  
During the forward pass it acts as an identity,
\[
\widetilde{z}_{\mathrm{txt}}^{(k)}=\mathcal{R}\!\left(z_{\mathrm{txt}}^{(k)}\right)=z_{\mathrm{txt}}^{(k)},
\]
whereas in the backward pass it multiplies the gradients by \(-\lambda_{\mathrm{grl}}\) with \(\lambda_{\mathrm{grl}}>0\):
\begin{equation}
\frac{\partial\mathcal{L}}{\partial z_{\mathrm{txt}}^{(k)}}
   =-\lambda_{\mathrm{grl}}
     \frac{\partial\mathcal{L}}{\partial \widetilde{z}_{\mathrm{txt}}^{(k)}}.
\end{equation}
Consequently the discriminator minimises \(\mathcal{L}_{\mathrm{pair}}\) while the encoder is forced to maximise it, so the encoder learns to retain information that supports correct gene–value pairing. 

\subsection{Objective Functions} \label{method:objective}

During training the model minimises four complementary objectives.  The InfoNCE loss aligns paired image and text embeddings in the shared contrastive space.  The observable loss \(\mathcal{L}_{\mathrm{obs}}\) and the unobservable loss \(\mathcal{L}_{\mathrm{unobs}}\) guide Mask Feature Modeling.  The pair-aware adversarial loss \(\mathcal{L}_{\mathrm{pair}}\) encourages the Gene-Text Encoder to encode only legitimate gene–value relations.  The combined loss is
\begin{equation}
\mathcal{L}= \mathcal{L}_{\mathrm{InfoNCE}} + \lambda_{1}\bigl(\mathcal{L}_{\mathrm{obs}} + \mathcal{L}_{\mathrm{unobs}}\bigr) + \lambda_{2}\mathcal{L}_{\mathrm{pair}},
\end{equation}
where \(\lambda_{1}\) and \(\lambda_{2}\) are weighting coefficients that balance reconstruction fidelity and pair-aware regularisation against the primary contrastive objective.  In all experiments we set \(\lambda_{1}=1\) and \(\lambda_{2}=0.1\).  This formulation jointly optimises cross-modal alignment, visual robustness, and pair-specific textual semantics, providing a unified training signal for end-to-end learning.

\section{Experiments}

\subsection{Datasets and Implementation Details} \label{method:datasets}

\noindent\textbf{Dataset:} For the dataset, we use STimage-1K4M \cite{chen2024stimage}, a recently released large-scale benchmark that aggregates 1149 Visium and Visium-HD whole-slide H\&E images together with spatial transcriptomes measured at every capture spot.  Each slide is tessellated into sub-image tiles that are linked to a high-dimensional vector containing 15,000 to 30,000 gene expression values, which yields a total of 4,293,195 image–gene pairs.  The dataset covers brain, breast, prostate, lung, liver, and lymph node tissue and provides precise spot coordinates so that visual context can be matched with molecular readouts at cellular resolution. 

\noindent\textbf{Implementation Details:} The vision backbone \(\text{ViT‑B/32}\) together with the image and text projection heads is the same as PLIP \cite{zuo2024plip}. Training is conducted on 8 NVIDIA RTX 4090 GPUs. The network is trained for 15 epochs. All model parameters are updated with the AdamW optimiser whose configuration is weight decay \(0.05\), learning rate \(1\times10^{-6}\) with a cosine decay schedule. 

\begin{table*}[t]
\centering
\caption{Adjusted Rand index (ARI) obtained by each model on the five evaluation brain slides 151672–151676.  Higher values indicate closer agreement between predicted clusters and the ground-truth cortical layer annotations.}
\label{tab:ari_slices}
\begin{tabular}{lcccccc}
\toprule
Method & 151672 & 151673 & 151674 & 151676 & 151675 & Overall \\
\midrule
CLIP               & 0.17 & 0.21 & 0.22 & 0.16 & 0.17 & 0.19 \\
PLIP               & 0.19 & 0.23 & 0.24 & 0.21 & 0.20 & 0.21 \\
UNI                & 0.19 & 0.21 & 0.21 & 0.18 & 0.17 & 0.19 \\
STimage\textendash1K4M & 0.23 & 0.25 & 0.22 & 0.26 & 0.22 & 0.24 \\
OmiCLIP            & 0.26 & 0.28 & 0.27 & 0.25 & 0.25 & 0.26 \\
\textbf{CoMTIP}              & \textbf{0.31} & \textbf{0.37} & \textbf{0.34} & \textbf{0.32} & \textbf{0.30} & \textbf{0.33} \\
\bottomrule
\end{tabular}
\end{table*}

\subsection{Tasks}

For evaluation we use a held‑out set of 18\,379 patch–gene pairs drawn from five human‑brain Visium slides.  CoMTIP is pre-trained on every remaining human slide in STimage-1K4M. After pre-training, we examine the model on two downstream tasks that probe complementary aspects of its transfer ability.
 
\noindent\textbf{Task 1 - Spatial clustering.} This unsupervised task evaluates zero-shot spatial understanding.  The five held-out brain slides are used for testing.  The slide provides spot-level RNA profiles with ground truth cortical layer annotations.  Every spot image is passed through the frozen image encoder to obtain a fused embedding.  K-means clustering is performed in this embedding space, and the resulting clusters are compared with the cortical layer labels.  

\noindent\textbf{Task 2 - Gene expression prediction.} This task quantifies the accuracy of regressing continuous expression values at individual spots.  Eight marker genes (IGKC, NPY, PLP1, HBB, SCGB2A2, MGP, GFAP, and MBP) are considered, and both evaluation modes use the same five brain slides.

\noindent\textbf{\textit{(1) Supervised mode}.} For each slide, 80\% of the spots are randomly selected for fine-tuning the entire CoMTIP network.  A lightweight multilayer perceptron is appended to the image encoder, and the whole architecture is optimised end to end to map image embeddings to the target gene values.  The remaining 20\% spots serve as the hold-out test split.

\noindent\textbf{\textit{(2) Zero-shot mode}.} Moreover, out CoMTIP has the zero-shot potential of CoMTIP for gene prediction. Predicted values are derived directly from similarity scores between the image embeddings and the template sentences, which reveals the inherent predictive capacity acquired during pre-training.  This protocol demonstrates the zero-shot potential of CoMTIP for gene prediction. The full procedure of using CoMTIP for zero-shot gene prediction are provided in the Supplementary Material.

We further evaluate the generalisation and effectiveness of our CoMTIP on external spatial transcriptomics datasets by performing the gene expression prediction task. The complete protocol, together with detailed results, is also reported in the \textit{Supplementary Materials}.

\begin{figure*}[t]
    \centering
    \includegraphics[width=\linewidth]{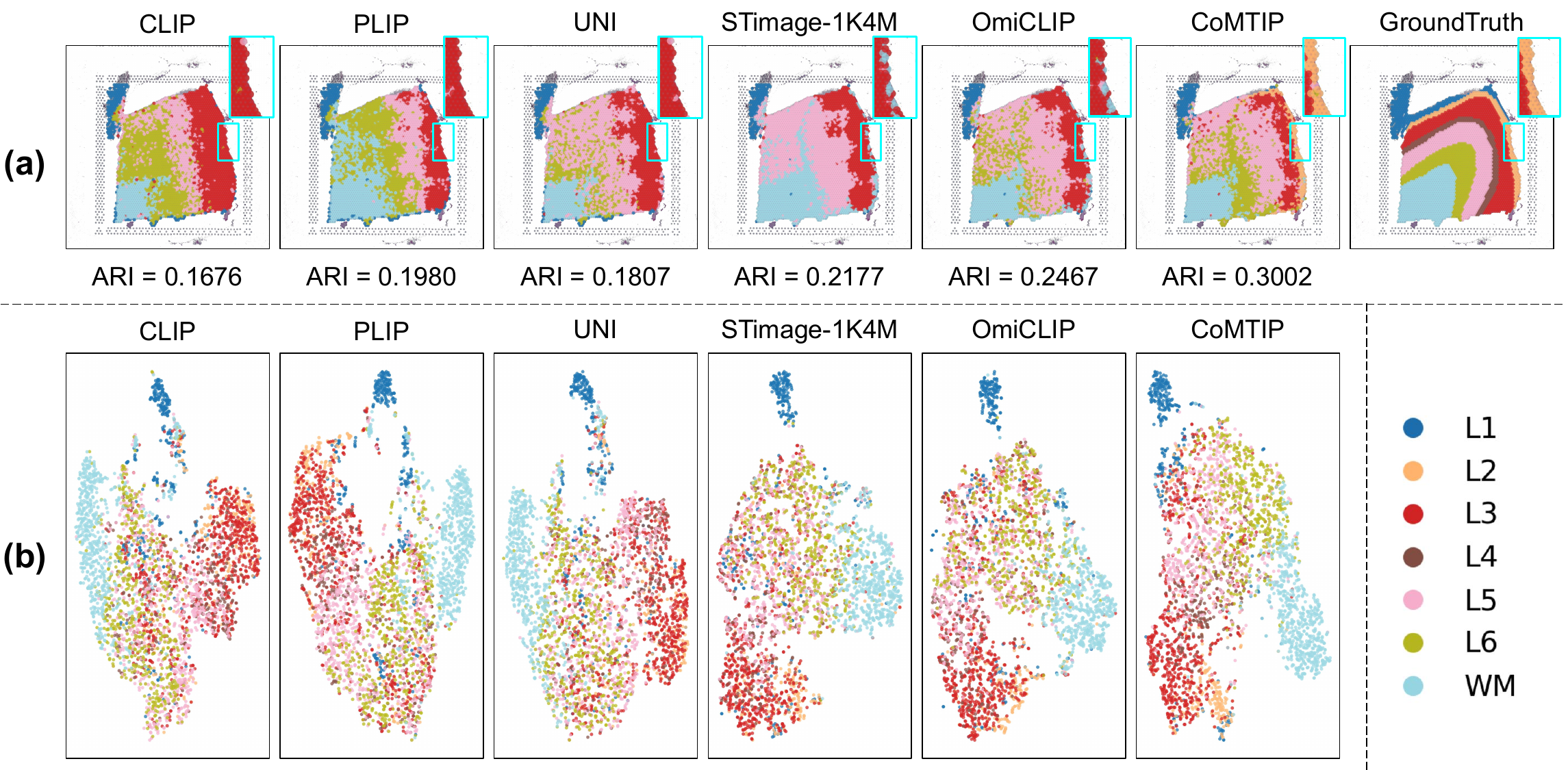}
    \caption{Unsupervised spatial clustering results on human-brain slide 151675.  (a) Ground-truth cortical layers compared with the layer assignments predicted by six models (CLIP, PLIP, UNI, STimage-1K4M, OmiCLIP and our CoMTIP) after clustering. (b) t-SNE visualisations of spot embeddings generated by different models. Each point is coloured by its ground-truth cortical label, revealing how clearly the embedding separates biological layers. CoMTIP provides the sharpest delineation among layers.}
    \label{fig:tsne}
\end{figure*}

\subsection{Baselines and Metrics}

\noindent\textbf{Baselines:} Five reference models are employed for comparison.  CLIP \cite{radford2021learning} is the original image-to-text contrastive model, and it is fine-tuned end-to-end on the training split.  PLIP \cite{zuo2024plip} augments CLIP with additional pathology specific pre-training which improves visual grounding in histology.  UNI \cite{chen2024towards} is a unified vision–language model that mixes natural and medical images during pre-training and is adapted to the current task with the same fine-tuning protocol.  STimage-1K4M \cite{chen2024stimage} aligns a small subset of gene values with visual features through contrastive learning and is retrained on the present data.  OmiCLIP \cite{chen2025visual} encodes gene names only and it too is fine-tuned under identical settings.  All five baselines therefore participate in the supervised experiment under the same data split as CoMTIP.


\noindent\textbf{Evaluation metrics:} For \textit{spatial clustering} (task 1), Adjusted Rand Index (ARI) measures the similarity between the clustering produced by the model and the ground truth cortical layer partition while correcting for random agreement, and a larger ARI indicates more accurate spatial grouping.  For \textit{gene expression prediction} (task 2), accuracy is summarized with mean absolute error and Pearson correlation.  Gene–wise mean absolute error (MAE\textsubscript{gene}) is first computed for each of the eight target genes as the average absolute difference between the predicted and the reference expression values across all spots.  These eight scores are then averaged to obtain the overall mean absolute error (MAE\textsubscript{overall}) that reflects the model’s aggregate precision.  Pearson correlation coefficient (PCC\textsubscript{overall}) is calculated on the concatenated vector that stacks the predicted expression values of all genes and all spots and measures the linear concordance with the corresponding reference vector.  Lower MAE and higher PCC indicate better performance.


\begin{table*}[t]
\centering
\caption{MAE for eight marker genes together with the overall MAE and overall PCC.  Results are averaged over the five held-out human-brain Visium slides that were excluded from pre-training. $^{*}$ indicates that the difference between CoMTIP and OmiCLIP on overall metrics is significant at $p<0.01$.}
\label{tab:gene_mae}
\begin{tabular}{lccccccccccc}
\toprule
\multirow{2.5}{*}{Method} & \multicolumn{8}{c}{MAE\textsubscript{gene}} & \multirow{2.5}{*}{MAE\textsubscript{overall}} & \multirow{2.5}{*}{PCC\textsubscript{overall}}\\
\cmidrule(lr){2-9}
 & IGKC & NPY & PLP1 & HBB & SCGB2A2 & MGP & GFAP & MBP \\
\midrule
CLIP               & 0.30 & 0.17 & 0.33 & 0.19 & 0.40 & 0.32 & 0.34 & 0.27 & 0.29 & 0.33 \\
PLIP               & 0.30 & 0.16 & 0.30 & 0.20 & 0.37 & 0.31 & 0.32 & 0.28 & 0.28 & 0.33 \\
UNI                & 0.32 & 0.15 & 0.35 & 0.22 & 0.41 & 0.34 & 0.33 & 0.32 & 0.31 & 0.30 \\
STimage\textendash1K4M & 0.28 & 0.13 & 0.29 & 0.19 & 0.38 & 0.30 & 0.31 & 0.24 & 0.26 & 0.36 \\
OmiCLIP            & 0.27 & 0.15 & 0.30 & \textbf{0.17} & 0.39 & 0.28 & 0.31 & 0.25 & 0.26 & 0.39 \\
\textbf{CoMTIP}              & \textbf{0.25} & \textbf{0.11} & \textbf{0.26} & \textbf{0.17} & \textbf{0.33} & \textbf{0.27} & \textbf{0.25} & \textbf{0.20} & \textbf{0.23$^{*}$} & \textbf{0.46$^{*}$} \\
\bottomrule
\end{tabular}
\end{table*}

\subsection{Spatial Clustering}

Table~\ref{tab:ari_slices} reports the ARI obtained by six models on the five held-out brain slides.  CoMTIP yields the highest score on every specimen with values ranging from \(0.30\) to \(0.37\), substantially surpassing the spatially oriented baseline OmiCLIP, whose ARI varies between \(0.25\) and \(0.28\).  STimage-1K4M provides the next best performance but remains consistently below CoMTIP by at least \(0.06\) absolute.  The generic vision–language models PLIP, UNI, and CLIP trail further behind and show greater variance across slides, which indicates limited robustness to anatomical heterogeneity.  The uniform superiority of CoMTIP confirms that its representations capture cortical structure more faithfully than both specialized and generic alternatives.

Figure~\ref{fig:tsne} offers a two-part evaluation of zero-shot spatial clustering on brain slide 151675 \cite{maynard2021transcriptome}.  The upper row juxtaposes the ground-truth cortical map with the predictions from six models.  CoMTIP delineates laminar boundaries more faithfully than competing models and uniquely identifies a contiguous band of L2 spots absent from all baselines.  Other models (CLIP, PLIP, UNI, STimage-1K4M, and OmiCLIP) misclassify large portions of the slice and get a lower ARI score.  The lower row visualises t-SNE projections of the spot embeddings generated by each model, with every point coloured by its ground-truth cortical label.  CoMTIP also outperforms the competing methods.  For instance, the t-SNE projection reveals that CoMTIP separates L2 from layer L6 more distinctly than any other layer combination.  These findings confirm that CoMTIP learns more discriminative spatial representations than both specialized and generic alternatives.

\subsection{Gene Expression Prediction}

The first block of Table~\ref{tab:gene_mae} reports the outcomes obtained after end-to-end finetuning on the 8 target genes.  CoMTIP reaches an overall MAE of \(0.23\) and an overall PCC of \(0.46\).  These scores improve on the strongest baseline STimage\,1K4M by \(0.03\) in error and \(0.10\) in correlation and surpass CLIP by \(0.06\) and \(0.13\) respectively.  CoMTIP delivers the best gene-wise error on seven of the eight genes and matches the leading result on the remaining two.  The largest gains appear for GFAP, where the error drops from \(0.31\) in STimage\,1K4M to \(0.25\), and for MBP, where the error decreases from \(0.24\) to \(0.20\).  These improvements suggest that the joint modeling of image content, gene identity, and expression magnitude provides richer features that generalize well across heterogeneous molecular patterns.

The last line presents the zero-shot variant that keeps all pretrained parameters frozen and infers expression with the template matching scheme.  Although the overall error rises to \(0.74\), the method still retains a correlation of \(0.21\) with no task-specific supervision.  This observation confirms that the shared embedding space encodes quantitative information that can be exploited without additional training, and it offers a cost-efficient alternative when labeled spots are scarce.

The supervised experiment in Table~\ref{tab:gene_mae} compares CoMTIP with two spatially oriented foundation models and three general vision–language baselines.  CoMTIP attains an overall MAE of \(0.23\) and an overall PCC coefficient of \(0.46\).  Compared to STimage\,1K4M, and OmiCLIP, which are both pre-trained specifically for spatial transcriptomics, CoMTIP lowers the error by \(0.03\) and raises the correlation by \(0.10\) and \(0.07\), respectively.  When contrasted with the generic models CLIP, PLIP, and UNI, CoMTIP reduces the error by up to \(0.08\) and improves the correlation by as much as \(0.16\).  On a per-gene basis CoMTIP consistently achieves lower prediction error than all competing models. These gains indicate that CoMTIP encodes richer gene expression cues than either the spatially oriented baselines or the generic vision–language models, which enables more effective fine-tuning and consistently stronger performance across diverse genes.

In the zero-shot setting, CoMTIP keeps all parameters frozen and predicts expression by template matching.  The overall MAE increases to \(0.74\) yet a PCC of \(0.21\) is preserved, which demonstrates that the pretrained embedding space already encodes informative quantitative cues.  This ability to provide meaningful estimates without any downstream fine-tuning offers a cost-effective avenue for exploratory gene profiling when annotated data are scarce.

\subsection{Ablation Study}

\begin{table}[t]
\centering
\setlength{\tabcolsep}{3pt}
\caption{Ablation study on gene expression prediction.  The table reports overall mean absolute error (MAE\textsubscript{overall}) and Pearson correlation coefficient (PCC\textsubscript{overall}) for the full model and four ablated variants.}
\label{tab:ablation}
\begin{tabular}{lcc}
\toprule
Method & MAE\textsubscript{overall} & PCC\textsubscript{overall}\\
\midrule
\textbf{Full CoMTIP} & \textbf{0.23} & \textbf{0.46} \\
w/o Mask Feature Modeling & 0.24 & 0.43 \\
w/o PAAT & 0.25 & 0.40 \\
w/o Gene-name Embedding & 0.24 & 0.41 \\
w/o Value-Embedding & 0.25 & 0.39 \\
\bottomrule
\end{tabular}
\end{table}

Table~\ref{tab:ablation} reports the overall mean absolute error and Pearson correlation coefficient for the full model and four ablated variants.  Removing Masked Feature Modeling increases the error from \(0.23\) to \(0.24\) and lowers the correlation from \(0.46\) to \(0.43\), suggesting that reconstructing occluded patches contributes to more precise visual features.  When PAAT is disabled, the error rises further to \(0.25\) and the correlation drops to \(0.40\), indicating that the adversarial signal is important for preserving accurate gene–value semantics.  Eliminating the gene-name embedding deteriorates performance to \(0.24\) MAE and \(0.41\) PCC, while removing the value embedding produces the largest degradation with \(0.25\) MAE and \(0.39\) PCC.  These results indicate that both categorical and numerical embeddings are essential for capturing the full spectrum of molecular information and that each component of CoMTIP makes a measurable contribution to the final accuracy.

\section{Discussion}

CoMTIP demonstrates that coupling context-aware visual reconstruction with pair-aware textual modeling yields a versatile foundation model for spatial transcriptomics.  It aligns whole-slide imagery with genome-scale transcriptomes in a single latent space, supports zero-shot molecular inference, and attains state-of-the-art accuracy after minimal fine-tuning. 

However, our proposed method leaves room for future improvement.  Because CoMTIP inherits the CLIP-style architecture, where the text encoder is capped at seventy-seven tokens, each transcriptome must be split into many short sentences, and this fragmentation inflates computation.  Adopting encoders with longer context windows could remove this bottleneck and permit richer gene narratives \cite{alayrac2022flamingo, li2023blip, chen2023pali, liu2023visual, wang2024qwen2}.  Future work may therefore integrate CoMTIP with such models to handle entire transcriptomes in a single forward pass and to reduce latency.  In addition, the fixed prompt “The expression value of gene is value” likely under-represents biological nuance.  Prompt-learning techniques, including soft prompts, prefix tuning, and instruction tuning \cite{zhou2022conditional, zhou2022learning, shen2024multitask} could adapt the template to tissue type or sequencing platform and further enhance accuracy.  Exploring these directions may unlock even broader utility for spatial transcriptomics studies.

\section{Conclusion}

We have proposed CoMTIP, a Contrastive Masked Text-Image Pretraining framework for spatial transcriptomics that jointly encodes whole-slide histology and genome-scale transcriptomes.  Its vision branch employs Masked-Feature Modeling that reconstructs occluded patches and thus learns robust context-aware representations for images.  The text branch introduces a scalable Gene-Text Encoder that assigns dedicated embeddings to each gene name and its expression value, allowing the model to process lots of gene–value pairs in a single forward pass. Pair-Aware Adversarial Training further regularises this branch by exposing the encoder to mismatched sentences so that only information supportive of correct gene–value pairing is retained.  By projecting both modalities into a shared latent space, CoMTIP achieves principled alignment of morphological context and molecular magnitude, which in turn and supplies a strong starting point for diverse downstream analyses in spatial transcriptomics.

\bibliography{aaai25}

\begin{thebibliography}{40}
\providecommand{\natexlab}[1]{#1}

\bibitem[{Alayrac et~al.(2022)Alayrac, Donahue, Luc, Miech, Barr, Hasson, Lenc, Mensch, Millican, Reynolds et~al.}]{alayrac2022flamingo}
Alayrac, J.-B.; Donahue, J.; Luc, P.; Miech, A.; Barr, I.; Hasson, Y.; Lenc, K.; Mensch, A.; Millican, K.; Reynolds, M.; et~al. 2022.
\newblock Flamingo: a visual language model for few-shot learning.
\newblock \emph{Advances in neural information processing systems}, 35: 23716--23736.

\bibitem[{Chen et~al.(2022)Chen, Liao, Cheng, Ma, Wu, Lai, Qiu, Yang, Xu, Hao et~al.}]{chen2022spatiotemporal}
Chen, A.; Liao, S.; Cheng, M.; Ma, K.; Wu, L.; Lai, Y.; Qiu, X.; Yang, J.; Xu, J.; Hao, S.; et~al. 2022.
\newblock Spatiotemporal transcriptomic atlas of mouse organogenesis using DNA nanoball-patterned arrays.
\newblock \emph{Cell}, 185(10): 1777--1792.

\bibitem[{Chen et~al.(2024{\natexlab{a}})Chen, Zhou, Wu, Zhang, Li, and Li}]{chen2024stimage}
Chen, J.; Zhou, M.; Wu, W.; Zhang, J.; Li, Y.; and Li, D. 2024{\natexlab{a}}.
\newblock STimage-1K4M: A histopathology image-gene expression dataset for spatial transcriptomics.
\newblock \emph{ArXiv}, arXiv--2406.

\bibitem[{Chen et~al.(2024{\natexlab{b}})Chen, Ding, Lu, Williamson, Jaume, Song, Chen, Zhang, Shao, Shaban et~al.}]{chen2024towards}
Chen, R.~J.; Ding, T.; Lu, M.~Y.; Williamson, D.~F.; Jaume, G.; Song, A.~H.; Chen, B.; Zhang, A.; Shao, D.; Shaban, M.; et~al. 2024{\natexlab{b}}.
\newblock Towards a general-purpose foundation model for computational pathology.
\newblock \emph{Nature Medicine}, 30(3): 850--862.

\bibitem[{Chen et~al.(2025)Chen, Zhang, Tran, Xiao, Li, Shah, Cheng, Brannan, Youker, Lai et~al.}]{chen2025visual}
Chen, W.; Zhang, P.; Tran, T.~N.; Xiao, Y.; Li, S.; Shah, V.~V.; Cheng, H.; Brannan, K.~W.; Youker, K.; Lai, L.; et~al. 2025.
\newblock A visual--omics foundation model to bridge histopathology with spatial transcriptomics.
\newblock \emph{Nature Methods}, 1--15.

\bibitem[{Chen et~al.(2023)Chen, Wang, Beyer, Kolesnikov, Wu, Voigtlaender, Mustafa, Goodman, Alabdulmohsin, Padlewski et~al.}]{chen2023pali}
Chen, X.; Wang, X.; Beyer, L.; Kolesnikov, A.; Wu, J.; Voigtlaender, P.; Mustafa, B.; Goodman, S.; Alabdulmohsin, I.; Padlewski, P.; et~al. 2023.
\newblock Pali-3 vision language models: Smaller, faster, stronger.
\newblock \emph{arXiv preprint arXiv:2310.09199}.

\bibitem[{Chung et~al.(2024)Chung, Ha, Im, and Lee}]{chung2024accurate}
Chung, Y.; Ha, J.~H.; Im, K.~C.; and Lee, J.~S. 2024.
\newblock Accurate spatial gene expression prediction by integrating multi-resolution features.
\newblock In \emph{Proceedings of the IEEE/CVF Conference on Computer Vision and Pattern Recognition}, 11591--11600.

\bibitem[{Ganguly et~al.(2025)Ganguly, Chatterjee, Huang, Zhang, Yurovsky, Johnson, and Chen}]{ganguly2025merge}
Ganguly, A.; Chatterjee, D.; Huang, W.; Zhang, J.; Yurovsky, A.; Johnson, T.~S.; and Chen, C. 2025.
\newblock MERGE: Multi-faceted Hierarchical Graph-based GNN for Gene Expression Prediction from Whole Slide Histopathology Images.
\newblock In \emph{Proceedings of the Computer Vision and Pattern Recognition Conference}, 15611--15620.

\bibitem[{Gong et~al.(2024)Gong, Arbesfeld-Qiu, Perrault, Bae, and Hwang}]{gong2024spatial}
Gong, D.; Arbesfeld-Qiu, J.~M.; Perrault, E.; Bae, J.~W.; and Hwang, W.~L. 2024.
\newblock Spatial oncology: Translating contextual biology to the clinic.
\newblock \emph{Cancer Cell}, 42(10): 1653--1675.

\bibitem[{He et~al.(2020{\natexlab{a}})He, Bergenstr{\aa}hle, Stenbeck, Abid, Andersson, Borg, Maaskola, Lundeberg, and Zou}]{he2020integrating}
He, B.; Bergenstr{\aa}hle, L.; Stenbeck, L.; Abid, A.; Andersson, A.; Borg, {\AA}.; Maaskola, J.; Lundeberg, J.; and Zou, J. 2020{\natexlab{a}}.
\newblock Integrating spatial gene expression and breast tumour morphology via deep learning.
\newblock \emph{Nature biomedical engineering}, 4(8): 827--834.

\bibitem[{He et~al.(2022)He, Chen, Xie, Li, Doll{\'a}r, and Girshick}]{he2022masked}
He, K.; Chen, X.; Xie, S.; Li, Y.; Doll{\'a}r, P.; and Girshick, R. 2022.
\newblock Masked autoencoders are scalable vision learners.
\newblock In \emph{Proceedings of the IEEE/CVF conference on computer vision and pattern recognition}, 16000--16009.

\bibitem[{He et~al.(2020{\natexlab{b}})He, Fan, Wu, Xie, and Girshick}]{he2020momentum}
He, K.; Fan, H.; Wu, Y.; Xie, S.; and Girshick, R. 2020{\natexlab{b}}.
\newblock Momentum contrast for unsupervised visual representation learning.
\newblock In \emph{Proceedings of the IEEE/CVF conference on computer vision and pattern recognition}, 9729--9738.

\bibitem[{Jia et~al.(2021)Jia, Yang, Xia, Chen, Parekh, Pham, Le, Sung, Li, and Duerig}]{jia2021scaling}
Jia, C.; Yang, Y.; Xia, Y.; Chen, Y.-T.; Parekh, Z.; Pham, H.; Le, Q.; Sung, Y.-H.; Li, Z.; and Duerig, T. 2021.
\newblock Scaling up visual and vision-language representation learning with noisy text supervision.
\newblock In \emph{International conference on machine learning}, 4904--4916. PMLR.

\bibitem[{Kuppe et~al.(2022)Kuppe, Ramirez~Flores, Li, Hayat, Levinson, Liao, Hannani, Tanevski, W{\"u}nnemann, Nagai et~al.}]{kuppe2022spatial}
Kuppe, C.; Ramirez~Flores, R.~O.; Li, Z.; Hayat, S.; Levinson, R.~T.; Liao, X.; Hannani, M.~T.; Tanevski, J.; W{\"u}nnemann, F.; Nagai, J.~S.; et~al. 2022.
\newblock Spatial multi-omic map of human myocardial infarction.
\newblock \emph{Nature}, 608(7924): 766--777.

\bibitem[{Li et~al.(2023)Li, Li, Savarese, and Hoi}]{li2023blip}
Li, J.; Li, D.; Savarese, S.; and Hoi, S. 2023.
\newblock Blip-2: Bootstrapping language-image pre-training with frozen image encoders and large language models.
\newblock In \emph{International conference on machine learning}, 19730--19742. PMLR.

\bibitem[{Li et~al.(2022)Li, Li, Xiong, and Hoi}]{li2022blip}
Li, J.; Li, D.; Xiong, C.; and Hoi, S. 2022.
\newblock Blip: Bootstrapping language-image pre-training for unified vision-language understanding and generation.
\newblock In \emph{International conference on machine learning}, 12888--12900. PMLR.

\bibitem[{Li et~al.(2021)Li, Selvaraju, Gotmare, Joty, Xiong, and Hoi}]{li2021align}
Li, J.; Selvaraju, R.; Gotmare, A.; Joty, S.; Xiong, C.; and Hoi, S. C.~H. 2021.
\newblock Align before fuse: Vision and language representation learning with momentum distillation.
\newblock \emph{Advances in neural information processing systems}, 34: 9694--9705.

\bibitem[{Liu et~al.(2024)Liu, Zhao, Shen, Ding, and Deng}]{liu2024ressat}
Liu, A.; Zhao, Y.; Shen, H.; Ding, Z.; and Deng, H.-W. 2024.
\newblock ResSAT: Enhancing Spatial Transcriptomics Prediction from H\&E-Stained Histology Images with Interactive Spot Transformer.
\newblock \emph{Research Square}, rs--3.

\bibitem[{Liu et~al.(2023)Liu, Li, Wu, and Lee}]{liu2023visual}
Liu, H.; Li, C.; Wu, Q.; and Lee, Y.~J. 2023.
\newblock Visual instruction tuning.
\newblock \emph{Advances in neural information processing systems}, 36: 34892--34916.

\bibitem[{Maynard et~al.(2021)Maynard, Collado-Torres, Weber, Uytingco, Barry, Williams, Catallini, Tran, Besich, Tippani et~al.}]{maynard2021transcriptome}
Maynard, K.~R.; Collado-Torres, L.; Weber, L.~M.; Uytingco, C.; Barry, B.~K.; Williams, S.~R.; Catallini, J.~L.; Tran, M.~N.; Besich, Z.; Tippani, M.; et~al. 2021.
\newblock Transcriptome-scale spatial gene expression in the human dorsolateral prefrontal cortex.
\newblock \emph{Nature neuroscience}, 24(3): 425--436.

\bibitem[{Moncada et~al.(2020)Moncada, Barkley, Wagner, Chiodin, Devlin, Baron, Hajdu, Simeone, and Yanai}]{moncada2020integrating}
Moncada, R.; Barkley, D.; Wagner, F.; Chiodin, M.; Devlin, J.~C.; Baron, M.; Hajdu, C.~H.; Simeone, D.~M.; and Yanai, I. 2020.
\newblock Integrating microarray-based spatial transcriptomics and single-cell RNA-seq reveals tissue architecture in pancreatic ductal adenocarcinomas.
\newblock \emph{Nature biotechnology}, 38(3): 333--342.

\bibitem[{Mondol et~al.(2023)Mondol, Millar, Graham, Browne, Sowmya, and Meijering}]{mondol2023hist2rna}
Mondol, R.~K.; Millar, E.~K.; Graham, P.~H.; Browne, L.; Sowmya, A.; and Meijering, E. 2023.
\newblock hist2rna: an efficient deep learning architecture to predict gene expression from breast cancer histopathology images.
\newblock \emph{Cancers}, 15(9): 2569.

\bibitem[{Radford et~al.(2021)Radford, Kim, Hallacy, Ramesh, Goh, Agarwal, Sastry, Askell, Mishkin, Clark et~al.}]{radford2021learning}
Radford, A.; Kim, J.~W.; Hallacy, C.; Ramesh, A.; Goh, G.; Agarwal, S.; Sastry, G.; Askell, A.; Mishkin, P.; Clark, J.; et~al. 2021.
\newblock Learning transferable visual models from natural language supervision.
\newblock In \emph{International conference on machine learning}, 8748--8763. PmLR.

\bibitem[{Rodriques et~al.(2019)Rodriques, Stickels, Goeva, Martin, Murray, Vanderburg, Welch, Chen, Chen, and Macosko}]{rodriques2019slide}
Rodriques, S.~G.; Stickels, R.~R.; Goeva, A.; Martin, C.~A.; Murray, E.; Vanderburg, C.~R.; Welch, J.; Chen, L.~M.; Chen, F.; and Macosko, E.~Z. 2019.
\newblock Slide-seq: A scalable technology for measuring genome-wide expression at high spatial resolution.
\newblock \emph{Science}, 363(6434): 1463--1467.

\bibitem[{Schmauch et~al.(2020)Schmauch, Romagnoni, Pronier, Saillard, Maill{\'e}, Calderaro, Kamoun, Sefta, Toldo, Zaslavskiy et~al.}]{schmauch2020deep}
Schmauch, B.; Romagnoni, A.; Pronier, E.; Saillard, C.; Maill{\'e}, P.; Calderaro, J.; Kamoun, A.; Sefta, M.; Toldo, S.; Zaslavskiy, M.; et~al. 2020.
\newblock A deep learning model to predict RNA-Seq expression of tumours from whole slide images.
\newblock \emph{Nature communications}, 11(1): 3877.

\bibitem[{Shen et~al.(2024)Shen, Yang, Zhang, Zhai, Gonzalez, Keutzer, and Darrell}]{shen2024multitask}
Shen, S.; Yang, S.; Zhang, T.; Zhai, B.; Gonzalez, J.~E.; Keutzer, K.; and Darrell, T. 2024.
\newblock Multitask vision-language prompt tuning.
\newblock In \emph{Proceedings of the IEEE/CVF Winter Conference on Applications of Computer Vision}, 5656--5667.

\bibitem[{Song et~al.(2024)Song, Cosatto, Wang, Kuang, Gerstein, Min, and Warrell}]{song2024predicting}
Song, T.; Cosatto, E.; Wang, G.; Kuang, R.; Gerstein, M.; Min, M.~R.; and Warrell, J. 2024.
\newblock Predicting spatially resolved gene expression via tissue morphology using adaptive spatial GNNs.
\newblock \emph{Bioinformatics}, 40(Supplement\_2): ii111--ii119.

\bibitem[{St{\aa}hl et~al.(2016)St{\aa}hl, Salm{\'e}n, Vickovic, Lundmark, Navarro, Magnusson, Giacomello, Asp, Westholm, Huss et~al.}]{staahl2016visualization}
St{\aa}hl, P.~L.; Salm{\'e}n, F.; Vickovic, S.; Lundmark, A.; Navarro, J.~F.; Magnusson, J.; Giacomello, S.; Asp, M.; Westholm, J.~O.; Huss, M.; et~al. 2016.
\newblock Visualization and analysis of gene expression in tissue sections by spatial transcriptomics.
\newblock \emph{Science}, 353(6294): 78--82.

\bibitem[{Stickels et~al.(2021)Stickels, Murray, Kumar, Li, Marshall, Di~Bella, Arlotta, Macosko, and Chen}]{stickels2021highly}
Stickels, R.~R.; Murray, E.; Kumar, P.; Li, J.; Marshall, J.~L.; Di~Bella, D.~J.; Arlotta, P.; Macosko, E.~Z.; and Chen, F. 2021.
\newblock Highly sensitive spatial transcriptomics at near-cellular resolution with Slide-seqV2.
\newblock \emph{Nature biotechnology}, 39(3): 313--319.

\bibitem[{Vickovic et~al.(2019)Vickovic, Eraslan, Salm{\'e}n, Klughammer, Stenbeck, Schapiro, {\"A}ij{\"o}, Bonneau, Bergenstr{\aa}hle, Navarro et~al.}]{vickovic2019high}
Vickovic, S.; Eraslan, G.; Salm{\'e}n, F.; Klughammer, J.; Stenbeck, L.; Schapiro, D.; {\"A}ij{\"o}, T.; Bonneau, R.; Bergenstr{\aa}hle, L.; Navarro, J.~F.; et~al. 2019.
\newblock High-definition spatial transcriptomics for in situ tissue profiling.
\newblock \emph{Nature methods}, 16(10): 987--990.

\bibitem[{Wang et~al.(2024)Wang, Bai, Tan, Wang, Fan, Bai, Chen, Liu, Wang, Ge et~al.}]{wang2024qwen2}
Wang, P.; Bai, S.; Tan, S.; Wang, S.; Fan, Z.; Bai, J.; Chen, K.; Liu, X.; Wang, J.; Ge, W.; et~al. 2024.
\newblock Qwen2-vl: Enhancing vision-language model's perception of the world at any resolution.
\newblock \emph{arXiv preprint arXiv:2409.12191}.

\bibitem[{Wang et~al.(2022)Wang, Wu, Agarwal, and Sun}]{wang2022medclip}
Wang, Z.; Wu, Z.; Agarwal, D.; and Sun, J. 2022.
\newblock Medclip: Contrastive learning from unpaired medical images and text.
\newblock In \emph{Proceedings of the Conference on Empirical Methods in Natural Language Processing. Conference on Empirical Methods in Natural Language Processing}, volume 2022, 3876.

\bibitem[{Wang et~al.(2021)Wang, Yu, Yu, Dai, Tsvetkov, and Cao}]{wang2021simvlm}
Wang, Z.; Yu, J.; Yu, A.~W.; Dai, Z.; Tsvetkov, Y.; and Cao, Y. 2021.
\newblock Simvlm: Simple visual language model pretraining with weak supervision.
\newblock \emph{arXiv preprint arXiv:2108.10904}.

\bibitem[{Xie et~al.(2023)Xie, Pang, Chung, Perciani, MacParland, Wang, and Bader}]{xie2023spatially}
Xie, R.; Pang, K.; Chung, S.; Perciani, C.; MacParland, S.; Wang, B.; and Bader, G. 2023.
\newblock Spatially resolved gene expression prediction from histology images via bi-modal contrastive learning.
\newblock \emph{Advances in Neural Information Processing Systems}, 36: 70626--70637.

\bibitem[{Xie et~al.(2022)Xie, Zhang, Cao, Lin, Bao, Yao, Dai, and Hu}]{xie2022simmim}
Xie, Z.; Zhang, Z.; Cao, Y.; Lin, Y.; Bao, J.; Yao, Z.; Dai, Q.; and Hu, H. 2022.
\newblock Simmim: A simple framework for masked image modeling.
\newblock In \emph{Proceedings of the IEEE/CVF conference on computer vision and pattern recognition}, 9653--9663.

\bibitem[{Zheng et~al.(2024)Zheng, Pizurica, Carrillo-Perez, Noor, Yao, Wohlfart, Marchal, Vladimirova, and Gevaert}]{zheng2024digital}
Zheng, Y.; Pizurica, M.; Carrillo-Perez, F.; Noor, H.; Yao, W.; Wohlfart, C.; Marchal, K.; Vladimirova, A.; and Gevaert, O. 2024.
\newblock Digital profiling of cancer transcriptomes from histology images with grouped vision attention.
\newblock \emph{BioRxiv}, 2023--09.

\bibitem[{Zhou et~al.(2022{\natexlab{a}})Zhou, Yang, Loy, and Liu}]{zhou2022conditional}
Zhou, K.; Yang, J.; Loy, C.~C.; and Liu, Z. 2022{\natexlab{a}}.
\newblock Conditional prompt learning for vision-language models.
\newblock In \emph{Proceedings of the IEEE/CVF conference on computer vision and pattern recognition}, 16816--16825.

\bibitem[{Zhou et~al.(2022{\natexlab{b}})Zhou, Yang, Loy, and Liu}]{zhou2022learning}
Zhou, K.; Yang, J.; Loy, C.~C.; and Liu, Z. 2022{\natexlab{b}}.
\newblock Learning to prompt for vision-language models.
\newblock \emph{International Journal of Computer Vision}, 130(9): 2337--2348.

\bibitem[{Zhu et~al.(2025)Zhu, Li, Tang, and Chang}]{zhu2025dusted}
Zhu, J.; Li, Y.; Tang, Z.; and Chang, C. 2025.
\newblock DUSTED: Dual-Attention Enhanced Spatial Transcriptomics Denoiser.
\newblock In \emph{Proceedings of the AAAI Conference on Artificial Intelligence}, volume~39, 1219--1227.

\bibitem[{Zuo et~al.(2024)Zuo, Hong, Zhang, Yu, Zhou, Gao, Sang, and Wang}]{zuo2024plip}
Zuo, J.; Hong, J.; Zhang, F.; Yu, C.; Zhou, H.; Gao, C.; Sang, N.; and Wang, J. 2024.
\newblock Plip: Language-image pre-training for person representation learning.
\newblock \emph{Advances in Neural Information Processing Systems}, 37: 45666--45702.

\end{thebibliography}

\end{document}


\maketitle

\section{Detailed Related Works}

\subsection{Spatial Transcriptomics}

Spatial transcriptomics has evolved from early array layouts that yielded near-cellular mRNA maps to bead-based platforms capable of genome-wide profiling at fine resolution \cite{vickovic2019high, rodriques2019slide, stickels2021highly, kuppe2022spatial}. Parallel to these advances, computational pipelines have been developed to infer expression directly from haematoxylin-and-eosin images. Initial convolutional models predicted bulk RNA-seq signals \cite{schmauch2020deep}. Subsequent work incorporated graph attention \cite{song2024predicting}, cross-modal contrastive learning \cite{xie2023spatially}, and multi-resolution fusion \cite{chung2024accurate}. Transformer architectures further improved accuracy by modelling spot neighbourhoods \cite{liu2024ressat} or denoising spatial domains \cite{zhu2025dusted}. 

Most of these methods rely on single-cohort training and therefore generalise poorly across protocols. Large-scale efforts such as STimage-1K4M \cite{chen2024stimage} and Visual-Omics CLIP \cite{chen2025visual} address data breadth by pairing thousands of slides with transcriptomes, yet both still lack unified modelling of image content, gene names and expression values, leaving room for more comprehensive cross-modal frameworks.

\subsection{Vision–Language Pre-Training}

Contrastive vision–language pre-training aligns visual and textual modalities to learn transferable representations. CLIP demonstrated that large-scale image–caption pairs can yield robust visual encoders \cite{radford2021learning}. Subsequent extensions such as ALIGN \cite{jia2021scaling}, BLIP \cite{li2022blip}, and SIMVLM \cite{wang2021simvlm} improved scaling and multimodal fusion, while FLAMINGO introduced retrieval-augmented few-shot reasoning \cite{alayrac2022flamingo}. In medical applications, MEDCLIP targets radiology reports \cite{wang2022medclip} and PLIP enhances pathological image grounding \cite{zuo2024plip}. 

\begin{figure}[htbp]
    \centering
    \includegraphics[width=1\linewidth]{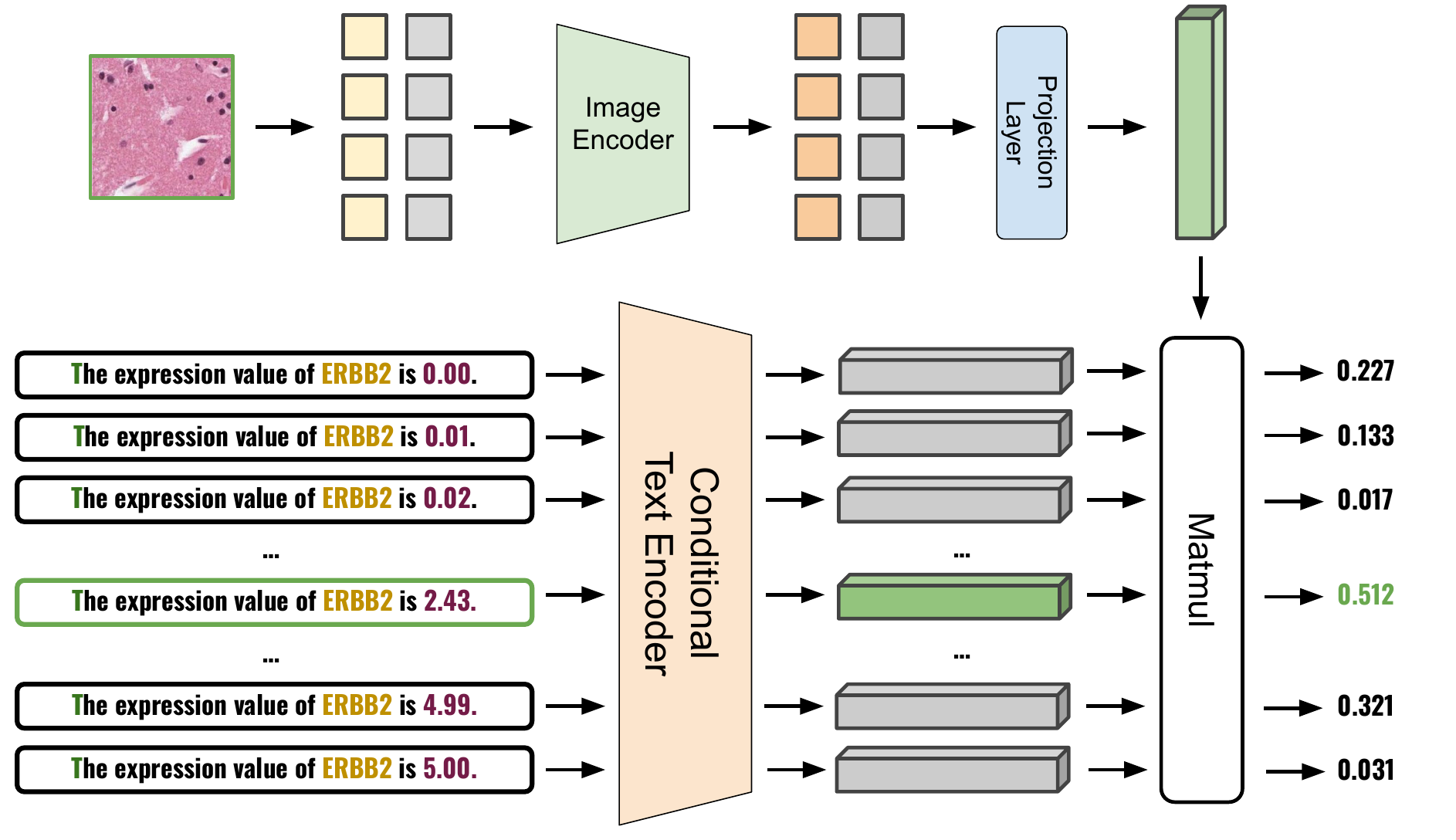}
    \caption{Zero-shot gene-expression estimation with CoMTIP.  A tissue image embedding is compared with embeddings of template sentences that encode candidate expression values, and the most similar template supplies the predicted magnitude.}
    \label{fig:5}
\end{figure}

\begin{table*}[ht]
\centering
\fontsize{7}{9}\selectfont
\setlength{\tabcolsep}{3pt}
\caption{Cross–validation performance on three Spatial Transcriptomics cohorts.  Metrics are reported as mean $\pm$ standard deviation across patient‑level folds.  PCC(M) denotes the mean Pearson correlation across all 250 genes, whereas PCC(H) is computed on the top‑50 highly predictive genes identified within each cohort.}
\label{tab:cv_results}
\begin{tabular}{lccc ccc ccc}
\toprule
\multirow{2.5}{*}{Source Model} &
\multicolumn{3}{c}{BC1} &
\multicolumn{3}{c}{BC2} &
\multicolumn{3}{c}{SCC} \\ \cmidrule(lr){2-4}\cmidrule(lr){5-7}\cmidrule(lr){8-10}
 & MSE & PCC(M) & PCC(H) & MSE & PCC(M) & PCC(H) & MSE & PCC(M) & PCC(H) \\
\midrule
CLIP \cite{radford2021learning}                          & 0.271 $\pm$ 0.06 & 0.236 $\pm$ 0.12 & 0.392 $\pm$ 0.15 & 0.225 $\pm$ 0.03 & 0.137 $\pm$ 0.07 & 0.282 $\pm$ 0.08 & 0.312 $\pm$ 0.11 & 0.298 $\pm$ 0.08 & 0.412 $\pm$ 0.09 \\
PLIP \cite{zuo2024plip}                             & 0.258 $\pm$ 0.05 & 0.262 $\pm$ 0.11 & 0.422 $\pm$ 0.14 & 0.219 $\pm$ 0.02 & 0.149 $\pm$ 0.06 & 0.294 $\pm$ 0.08 & 0.305 $\pm$ 0.10 & 0.318 $\pm$ 0.07 & 0.428 $\pm$ 0.08 \\
UNI \cite{chen2024towards}                               & 0.264 $\pm$ 0.07 & 0.228 $\pm$ 0.13 & 0.365 $\pm$ 0.16 & 0.234 $\pm$ 0.03 & 0.122 $\pm$ 0.06 & 0.266 $\pm$ 0.07 & 0.323 $\pm$ 0.12 & 0.304 $\pm$ 0.09 & 0.387 $\pm$ 0.10 \\
STimage‑1K4M \cite{chen2024stimage}                      & 0.247 $\pm$ 0.05 & 0.284 $\pm$ 0.13 & 0.453 $\pm$ 0.14 & 0.213 $\pm$ 0.03 & 0.171 $\pm$ 0.07 & 0.312 $\pm$ 0.09 & 0.283 $\pm$ 0.11 & 0.342 $\pm$ 0.08 & 0.456 $\pm$ 0.09 \\
OmiCLIP \cite{chen2025visual}                            & 0.240 $\pm$ 0.05 & 0.303 $\pm$ 0.12 & 0.483 $\pm$ 0.15 & 0.207 $\pm$ 0.03 & 0.192 $\pm$ 0.07 & 0.331 $\pm$ 0.09 & 0.285 $\pm$ 0.10 & 0.360 $\pm$ 0.07 & 0.477 $\pm$ 0.08 \\
\textbf{CoMTIP}                                          & \textbf{0.221 $\pm$ 0.06} & \textbf{0.365 $\pm$ 0.16} & \textbf{0.527 $\pm$ 0.13} & \textbf{0.200 $\pm$ 0.03} & \textbf{0.233 $\pm$ 0.06} & \textbf{0.337 $\pm$ 0.09} & \textbf{0.255 $\pm$ 0.10} & \textbf{0.401 $\pm$ 0.07} & \textbf{0.488 $\pm$ 0.06} \\
\bottomrule
\end{tabular}
\end{table*}

\section{Zero-shot Gene Expression Prediction} \label{method:zero}

As shown in Figure \ref{fig:5}, CoMTIP predicts gene expression without task-specific fine-tuning by directly exploiting the shared contrastive space learned during pre-training.  For a target gene \(g\) we first build a discrete template set
\[
\mathcal{T}_{g}= \bigl\{\,\text{'The expression value of } g \text{ is } v_{k}\text{.'}\,\bigr\}_{k=1}^{K},
\]
where \(\{v_{k}\}_{k=1}^{K}\) is an ordered list of candidate expression magnitudes obtained by uniform sampling over a biologically meaningful range.  Each template sentence is encoded by the Gene-Text Encoder to obtain vector representations
\[
z_{\mathrm{txt}}^{(k)} = P_{\mathrm{txt}}\!\bigl(E_{\mathrm{cond}}(\text{template } k)\bigr)\in\mathbb{R}^{d}.
\]

Given an unseen tissue image \(I\) its visual embedding is computed once
\[
z_{\mathrm{img}} = P_{\mathrm{img}}\!\bigl(E_{\mathrm{img}}(I)\bigr)\in\mathbb{R}^{d}.
\]
Similarity scores between the image and all templates are measured by normalized dot product
\begin{equation}
\sigma_{k}= \frac{\langle z_{\mathrm{img}},\,z_{\mathrm{txt}}^{(k)}\rangle}{\|z_{\mathrm{img}}\|_{2}\,\|z_{\mathrm{txt}}^{(k)}\|_{2}}, \qquad k=1,\dots,K.
\end{equation}

The hard selection rule assigns the expression estimate to the candidate value whose template attains the maximum similarity
\begin{equation}
\hat{v}_{g}= v_{k^{*}}, \quad k^{*}= \arg\max_{k}\,\sigma_{k}.
\end{equation}
Because both modalities share an aligned metric space the nearest template delivers an immediate quantitative prediction without additional parameters.  Increasing the density of \(\{v_{k}\}\) refines resolution at the cost of linear growth in template evaluations.  In practice \(K=501\) with values spaced at \(0.01\) intervals balances accuracy and computational overhead.

\section{Experiments on More Datasets} 

To ascertain the universality and generalization capacity of our pre-trained CoMTIP, we extended the evaluation to additional spatial transcriptomics datasets following precisely the patient‑level protocol gene panel selection and preprocessing pipeline defined by TRIPLEX \cite{chung2024accurate}. 

 Specifically, we evaluate our CoMTIP model on three cohorts spatial transcriptomics.  BC1 contains 36 breast cancer slides from 8 patients.  BC2 contains 68 breast cancer slides from 23 patients.  SCC contains 12 skin cancer slides from four patients.  Samples originating from the same patient are always assigned to a single partition in order to avoid information leakage.

For BC1 and SCC we employ leave‑one‑patient‑out cross‑validation.  Each fold reserves the slides of one patient for validation and uses the remaining patients for training.  For BC2 the larger sample size allows eight‑fold patient‑stratified cross‑validation where all slides of each patient remain in the same fold.

Following the ST‑Net protocol \cite{he2020integrating}, the top 250 genes ranked by mean expression in each cohort are selected as prediction targets and during cross‑validation the 50 highly predictive genes identified by average Pearson ranking are additionally reported. Spot count matrices are normalised by total count then transformed with a log base two function after adding a pseudo count of one and this expression is used without further scaling. 

Performance is quantified on held‑out patients by averaging the Pearson correlation coefficient across all genes denoted PCC‑A, the mean correlation across the highly predictive genes denoted PCC‑H, and absolute error metrics MAE. 

As shown in Table \ref{tab:cv_results}, CoMTIP attains the best performance on every metric across all three external cohorts, demonstrating strong generalisation beyond the data seen during pre‑training.  In BC1 it reduces the mean‑squared error to \(0.221\) and lifts the whole‑panel correlation to \(0.365\), surpassing the nearest competitor OmiCLIP by \(0.019\) MSE and \(0.062\) PCC.  Similar gains appear in BC2, where CoMTIP yields the lowest error of \(0.200\) and the highest full‑panel correlation of \(0.233\).  The advantage persists on the skin cancer cohort SCC, which differs in tissue type and experimental protocol, where CoMTIP improves the mean correlation to \(0.401\) while cutting the error to \(0.255\).  Performance on the highly predictive gene subset follows the same trend, confirming that the model captures both global and gene‑specific patterns.  These consistent improvements validate the universality of the learned representations and indicate that CoMTIP can transfer effectively to previously unseen spatial transcriptomics data.

\bibliography{aaai25}